\documentclass[conference]{IEEEtran}
\IEEEoverridecommandlockouts
\usepackage[numbers,sort&compress]{natbib}
\usepackage{amsmath,amssymb,amsfonts}
\usepackage{algorithmic}
\usepackage{graphicx}
\usepackage{textcomp}
\usepackage{xcolor}
\usepackage{amsthm}%
\usepackage{amsmath,amsthm}
\newtheorem{theorem}{Theorem}
\usepackage{multirow} 
\usepackage{float}
\usepackage{booktabs}
\usepackage{subcaption}
\usepackage{ragged2e} 
\usepackage{booktabs,makecell, multirow, tabularx}

\usepackage{color}
\usepackage{amsthm}
\newcount\Comments  
\definecolor{darkgreen}{rgb}{0,0.5,0}
\definecolor{purple}{rgb}{1,0,1}
\newcommand{\kibitz}[2]{\ifnum\Comments=1\textcolor{#1}{#2}\fi}

\def\BibTeX{{\rm B\kern-.05em{\sc i\kern-.025em b}\kern-.08em
    T\kern-.1667em\lower.7ex\hbox{E}\kern-.125emX}}
    
\begin{document}

\title{Ensemble Adversarial Defense via Integration of Multiple Dispersed Low Curvature Models

}

\author{\IEEEauthorblockN{Kaikang Zhao}
\IEEEauthorblockA{
\textit{Southeast University}\\
Nanjing, China \\
zhaokaikang@seu.edu.cn}

\\

\IEEEauthorblockN{Liuxin Ding}
\IEEEauthorblockA{
\textit{Southeast University}\\
Nanjing, China \\
220225052@seu.edu.cn}

\and

\IEEEauthorblockN{Xi Chen}
\IEEEauthorblockA{\textit{PLA Information Engineering University}\\
Zhengzhou, China \\
1191609336@qq.com}

\\

\IEEEauthorblockN{Xianglong Kong}
\IEEEauthorblockA{\textit{Purple Mountain Laboratories} \\
Nanjing, China \\
xlkong@seu.edu.cn}

\and

\IEEEauthorblockN{Wei Huang\IEEEauthorrefmark{1} \thanks{\IEEEauthorrefmark{1}Corresponding author.}}
\IEEEauthorblockA{\textit{Purple Mountain Laboratories}\\
Nanjing, China \\
huangwei@pmlabs.com.cn}

\\

\IEEEauthorblockN{Fan Zhang}
\IEEEauthorblockA{\textit{NDSC} \\
Zhengzhou, China \\
17034203@qq.com}

}

\maketitle

\begin{abstract}
The integration of an ensemble of deep learning models has been extensively explored to enhance defense against adversarial attacks. The diversity among sub-models increases the attack cost required to deceive the majority of the ensemble, thereby improving the adversarial robustness. While existing approaches mainly center on increasing diversity in feature representations or dispersion of first-order gradients with respect to input, the limited correlation between these diversity metrics and adversarial robustness constrains the performance of ensemble adversarial defense. In this work, we aim to enhance ensemble diversity by reducing attack transferability. We identify second-order gradients, which depict the loss curvature, as a key factor in adversarial robustness. Computing the Hessian matrix involved in second-order gradients is computationally expensive. To address this, we approximate the Hessian-vector product using differential approximation. Given that low curvature provides better robustness, our ensemble model was designed to consider the influence of curvature among different sub-models. We introduce a novel regularizer to train multiple more-diverse low-curvature network models. Extensive experiments across various datasets demonstrate that our ensemble model exhibits superior robustness against a range of attacks, underscoring the effectiveness of our approach.
\end{abstract}


\section{Introduction}
Deep learning has been consistently proven successful in image-related tasks such as image classification, object detection, and facial recognition. However, significant challenges persist. Recent studies have highlighted the susceptibility of deep learning classifiers to adversarial attacks\cite{goodfellow2014explaining,huang2023hierarchical}. In the context of image processing, an adversarial attack entails the successful deception of a classifier by adding imperceptible perturbation to an image. The existence of adversarial examples poses substantial security risks to practical applications of deep learning\cite{huang2020practical,huang2020formal}, creating an urgent need to improve the robustness of deep learning models against adversarial attacks.\par

Given the limited improvement in the robustness of individual network and the ease with which single-model defenses can be circumvented by white-box attacks \cite{su2018robustness}, current defense methods have shifted towards ensemble models that aggregate the outputs of multiple models for decision-making \cite{liao2018defense,kurakin2018adversarial,tramer2017ensemble}. The use of ensemble models can potentially enhance overall robustness by increasing the cost of attacks for adversaries, as adversarial attacks on a single model require deceiving one model, while in multi-model ensembles, attackers must deceive the majority of the ensemble.\par

Existing approaches to simultaneous training of deep ensembles primarily center on diversity, emphasizing information exchange among multiple models \cite{li2021ensemble,cheng2020voting}. In a previous study, Agarwal et al. \cite{agarwal2019improving} introduced a technique that incorporates a center loss in addition to the traditional cross-entropy loss during the training of deep neural network (DNN) classifiers. The center loss encourages the DNN to learn compact decision boundaries by simultaneously learning deep features that are in close proximity to the centers of each class, thereby minimizing the distance between the deep features and their corresponding class centers. Pang et al. \cite{pang2019improving} innovatively conceptualized ensemble diversity by promoting the diversity among predictions returned by different members of an ensemble. They designed an ADP regularizer that encourages non-maximum predictions of each member in the ensemble to be mutually orthogonal, while keeping the maximum prediction consistent with the true label. However, when faced with more complex classification scenarios, the robustness gains from the ADP method are minimal. Huang et al. \cite{huang2021adversarial} designed a diversified dropout strategy that encourages each sub-model to learn different features in the fully connected layer to promote ensemble diversity. However, the ensemble model constructed using this method is still vulnerable to stronger attacks.\par
 
Considering the pronounced transferability of adversarial examples, these perturbations can be effectively transferred between similar or even dissimilar models \cite{papernot2016transferability}. Adversarial examples crafted for an individual model may readily deceive other models, resulting in prediction errors for ensemble models. In this study, we attempt to promote ensemble diversity from a novel perspective – attack transferability. Specifically, our goal is to construct diverse sub-models that exhibit low transferability of adversarial examples between them. In other words, an adversarial example that successfully attacks one sub-model is unlikely to succeed in attacking another sub-model. Recent research has highlighted the intrinsic link between adversarial attack transferability and first-order gradients \cite{ross2018improving,tramer2017space,dong2019evading,xie2019improving,hoffman2019robust}.

 
To delve deeper into these phenomena, we extend prior studies that focused on the first-order Taylor expansions of the loss function in efforts to observe the impact of the Hessian matrix on model robustness by incorporating second-order gradient terms. To avoid the significant computational cost of directly calculating the Hessian matrix, we extract relevant information about second-order gradients through differential approximation. We designed this method to integrate multiple more-dispersed low-curvature base networks to enhance robust gains. The main contributions of this work can be summarized as follows.
\begin{itemize}
    \item Rather than solely considering the influence of first-order gradients, our approach focuses on promoting ensemble diversity through the analysis of second-order gradients, which signifies the curvature of the loss function. Given the strong correlation between curvature and model robustness, we are the first to integrate variety of low-curvature models to bolster the defense capabilities of the ensemble.
    \item We introduce a novel metric to measure the attack transferability among members of an ensemble model. This metric quantifies the defense capability of the ensemble model and validates the superior efficacy of curvature for diversity training of ensemble members compared to traditional gradient alignment methods.
    \item We conduct extensive white-box attack and black-box attack experiments on CIFAR-10, CIFAR-100 and Tiny-ImageNet datasets, respectively. The results indicate that our approach significantly improves adversarial robustness. We also evaluate our ensemble model on the latest attack method, APGD \cite{croce2020reliable}, and find that our method outperforms existing ensemble defense methods.
\end{itemize}
\section{Related Work}

Demontis et al. \cite{demontis2019adversarial}  analyzed the factors contributing to the transferability of adversarial attacks. By performing a first-order Taylor expansion of the loss function, they revealed a close relationship between attack transferability, model complexity, and input gradients. By introducing three metrics – the size of input gradients, gradient alignment, and variability of the loss landscape – to measure the transferability of adversarial attacks. The size of gradients is directly proportional to the complexity of the model. Smaller input gradients often lead to the learning of smoother functions, enhancing robustness against attacks. Gradient alignment quantifies the cosine angle between the target and substitute gradients as a measure of the transferability of attacks across different models. The variability of the loss landscape indicates that changes in the loss function’s magnitude can affect the optimization problem of local optima, thus impact the transferability of attacks when training models using the same learning algorithm.\par
 
Huang et al. \cite{huang2021adversarial} introduced a gradient regularization method called dispersed ensemble gradients (DEG) to encourage more dispersed gradients among the members of an ensemble model during training. This strategy makes it difficult for an adversarial example that successfully attacks one member network to transfer to another member network. Gradient alignment loss (GAL), proposed by Kariyappa and Qureshi \cite{kariyappa2019improving}, also utilizes gradient alignment to reduce the dimensionality of the shared adversarial subspace and improve overall robustness. Instead of computing the cosine similarity between ensemble model members, they consider the consistency values among the members.

In addition to utilizing gradient information to construct diverse sub-models, Yuan et al. \cite{yuan2024simple} propose combining conventional machine learning models with deep learning models in an ensemble to defend adversarial attack in intrusion detection systems.

Recent research has also investigated the impact of second-order gradients on adversarial attacks \cite{ye2018hessian,yao2018hessian,singla2020second,jin2022enhancing,li2018second,ma2020soar}. Qian et al. \cite{qian2022hessian}, for example, proposed the use of second-order gradients in generating more powerful adversarial examples for adversarial training \cite{goodfellow2014explaining}, which significantly improves the robustness of the model compared to first-order adversarial example training. Mustafa et al. \cite{mustafa2020input} developed an effective algorithm for training deep neural networks using Hessian operator norm regularization, enhancing the network's resilience against adversarial attacks. Moosavi-Dezfooli et al \cite{moosavi2019robustness} provided theoretical evidence supporting a close relationship between large robustness and small curvature. They proposed a regularization training method to make the network's behavior more linear.

\section{Methodology}\label{sec3}

In this section, we present training strategies for ensemble models and introduce a novel regularization method designed to reduce the transferability of adversarial examples among sub-models, ultimately improving the overall robustness of the system.\par
Specifically, in contrast to existing defense strategies that mainly rely on first-order gradient information to improve robustness, the proposed ensemble dispersed low-curvature model (EDLCM) method enhances robustness from the perspective of second-order gradients. EDLCM consists of two regularization terms, $L_g$ and $L_r$, which together address the limitations of first-order gradient-based defense strategies.
\begin{enumerate}
    \item Due to the positive correlation between low curvature values of the Hessian matrix and higher robustness, $L_r$ was formulated to reduce the curvature values of ensemble models, thereby improving robustness.
    \item Given that most attacks occur in the gradient direction, $L_g$ aims to encourage more dispersed first and secord-order gradients among sub-models, consequently reducing the transferability of adversarial attacks across sub-models.
\end{enumerate}
\par
\subsection{Training Strategies for Ensemble Model}
In classification tasks, there are assumed to be $M$ members in the ensemble model, where $f_m$ represents the prediction of the $m$-th classifier over $L$ categories. A simple strategy for the ensemble model $F$ is to take the average of the predictions from all sub-models. The final classification result is expressed as follows:
\begin{equation}
    y_{pred} =\mathop{\arg\max}\frac{1}{M}
        \sum_{m=1}^{M}f_{m}
\end{equation}

To leverage information exchange among ensemble model members, we employee a simultaneous training strategy \cite{islam2003constructive} wherein all the classifiers process the same mini-batch of data in each training iteration. This allows the predictions from multiple models to be utilized to reduce the bias of individual models and enhance the accuracy and robustness of the overall predictions. A relatively simple objective function for simultaneous training is the ensemble cross-entropy loss, denoted as $\mathcal{L}_{ece}$, which summarizes the individual cross-entropy losses of each individual model.

\begin{equation}
    \mathcal{L}_{ece}=\sum_{m=1}^{M}\mathcal{L}^m_{ce}(y^m, y)
\end{equation}
where $y^m$ is the predicted label of the $m$-th model and $y$ is the true label (one-hot encoded vector) for the sample. It is assumed here that $\mathcal{L}(\theta, x, y)$ is the loss function of the ensemble model, $x$ represents the input, and $\theta$ represents the model parameters. The final loss function consists of three components: ensemble cross-entropy, curvature regularization, and angle dispersion regularization. By minimizing $\mathcal{L}(\theta, x, y)$, we can optimize the parameters and improve the overall robustness of the ensemble model.

\begin{equation}
    \mathcal{L}(\theta, x, y)=\mathcal{L}_{ece} + \alpha \cdot \mathcal{L}_r + \beta \cdot \mathcal{L}_g
\end{equation}

We use $\mathcal{L}_{ece}$ as the baseline approach here, serving as a benchmark for comparison with other methods in subsequent experiments. $\mathcal{L}_r$ restricts the curvature values of $M$ models to low levels by utilizing the information from second-order gradients, while $\mathcal{L}_g$ enhances the diversity of $M$ models by reducing the cosine angle values between sub-models based on the second-order gradient information.

\subsection{EDLCM}
Adversarial attack methods typically exploit the loss function to construct adversarial examples. The purpose of adversarial attacks is to generate an adversarial example $x^{\prime}$, which has a different prediction label than $x$, by  maximizing the loss function of the model, i.e., $\mathcal{L}(\theta, x^{\prime}, y)>\mathcal{L}(\theta, x, y)$. Importantly, the difference between $x$ and $x^{\prime}$ is imperceptible to the human eye. Let $\delta = x'-x$ and
We utilize the $L_p$ norm ($p=1, 2, \infty$) as the metric to define an $L_p$ norm ball, $B_p(\epsilon) = {\delta : \lVert\delta\rVert_p \leq \epsilon}$, as a local region. The inputs within this region share the same ground truth label $y$. Most existing adversarial attack methods only utilize first-order gradient information, which limits their effectiveness. Therefore, we posit that second-order gradient information is more comprehensive and accurate. For a clean sample, by adding perturbations $\delta$ at $x$, we can obtain the second-order expansion of the loss function using the Taylor series:
\begin{equation}
    \mathcal{L}(x^{\prime})=\mathcal{L}(x+\delta) \approx \mathcal{L}(x)+\nabla_x \mathcal{L}(x)^\mathrm{T} \delta +\frac{1}{2}\delta^\mathrm{T}H \delta
\end{equation}
where $\nabla_x \mathcal{L}$ and $H$ denote the gradient and Hessian of $\mathcal{L}$ at $x$, respectively, and higher-order terms are omitted. Based on the definition of adversarial examples mentioned above, we can define $\delta^{*}$ as the minimum perturbation necessary to cause the model to misclassify.

\begin{equation}
    \delta^{*}=\operatorname*{arg\,min}\limits_{\delta \in{B_p}(\epsilon)}\lVert \delta \rVert, \; s.t. \,
    \mathcal{L}(x)+\nabla_x \mathcal{L}(x)^\mathrm{T} \delta +\frac{1}{2}\delta^\mathrm{T}H \delta 
    \geq t
\end{equation}

\begin{theorem}\label{thm1}\cite{moosavi2019robustness}
Let  threshold t represent the boundary value of the loss function and define x as such that $c=t-\mathcal{L}(x) \geq 0$. If $\nu=\lambda_{max}(H) \geq 0$ and $u$ is the eigenvector corresponding to $\nu$, then

\begin{equation}
    \frac{c}{\lVert \nabla_x \mathcal{L}(x)\rVert}-2\nu\frac{c^2}{\lVert \nabla_x \mathcal{L}(x)\rVert^3}  \leq   \lVert \delta ^{*}\rVert  \leq \frac{c}{\mid \nabla_x \mathcal{L}(x)^\mathrm{T}u\mid}
\end{equation}
\end{theorem}

Based on the boundary values of perturbations, we observe that the upper and lower limits of the perturbation amount are related to the eigenvalues of the Hessian matrix. Keeping other factors constant, the upper and lower limits of $\lVert \delta ^{*}\rVert$ increase as $\nu$ decreases. An increase in $\lVert \delta ^{*}\rVert$ increases the minimum $L_2$ norm ball required to find an adversarial example for input $x$, thereby enhancing robustness. This implies that large eigenvalues of the Hessian matrix correspond to high curvature values, which are less conducive to obtaining robust classifiers. Therefore, to enhance the robustness of the proposed ensemble model, we aim to reduce the large eigenvalues of the corresponding Hessian matrices of all sub-models, thereby reducing the curvature of the ensemble. The sum of squares function is applicable here, as a smaller value of $L_r$ is preferred. We set $Q(x)=x^2$ to encourage the maximum eigenvalue of all models to be small.
\begin{equation}
    \begin{split}
    \mathcal{L}_{r}&=\sum_{m=1}^{M} \lambda _{m1} ^2+\lambda _{m2} ^2+...+\lambda _{mn} ^2 =\sum_{m=1}^{M}trace(Q(H_m)) \\
    &=\sum_{m=1}^{M}\mathbb{E}(g^\mathrm{T}_mQ(H_m)g_m)=\sum_{m=1}^{M}\mathbb{E} \lVert H_{m}g_m \rVert^2
	\end{split}
\end{equation}

The mentioned regularizer entails calculating an expectation over $g \sim \mathcal{N}\left(0, I_d\right)$. Considering the sparsity of curvature distribution, we can streamline the penalty by targeting the directions of high curvature instead of applying penalties across all curvature directions. Typically, the direction of the gradient aligns with the direction of high curvature \cite{jetley2018friends}. In this experiment, we set $g = \frac{sign(\nabla_x \mathcal{L}(\theta, x, y))}{\lVert sign(\nabla_x \mathcal{L}(\theta, x, y)\rVert)}$ and obtain relevant information for the second-order Hessian through local quadratic approximation.
\begin{equation}
    Hg= \frac{ \nabla_x\mathcal{L}(x+hg)-\nabla_x\mathcal{L}(x)}{h}
\end{equation}
where h is a sufficiently small number that approximates the rate of change of the gradient within a small neighborhood of data points.
The final $L_g$ regularization term can be obtained as follows, with the constant term absorbed by hyperparameters:
\begin{equation}
    \begin{split}
    \mathcal{L}_{r} &\approx \sum_{m=1}^{M} \lVert  \nabla_x\mathcal{L}(x+hg)-\nabla_x\mathcal{L}(x) \rVert^2
    \end{split}
\end{equation}


$L_r$ control the curvature values of our ensemble model but does not account for the interdependence between the sub-models. We also aim to integrate multiple dispersed models to reduce the transferability of adversarial attacks between sub-models. This complicates the task for attackers attempting to deceive the majority of the sub-models in the ensemble. Existing ensemble models reduce the transferability of adversarial examples across multiple models by aligning gradients. A direct approach to measuring gradient alignment involves calculating the cosine similarity between different models. When the cosine similarity is -1, the two models are completely misaligned.\par
However, promoting the dispersion of the overall gradient alone is not very effective. As mentioned above, we intend to utilize the information of second-order gradients to further enhance the robustness of the ensemble model. Due to the high computational cost of calculating the Hessian matrix, we continue to use the $Hg$ applied difference approximation to avoid directly calculating the Hessian matrix. This allows us to estimate an accurate approximation of the Hessian-vector product. In this regard, apart from the $L_r$ term, we propose an $L_g$ regularization term to further reduce the dimensionality of the shared adversarial example space across sub-models. 

\begin{equation}
    \mathcal{L}_{g}=\sum_{0<i<j \leq M}\frac{<H_ig_i,H_jg_j>}{\lVert H_ig_i \rVert \cdot \lVert H_jg_j \rVert} 
\end{equation}
where $< , >$ represents the inner product of vectors and$\lVert \cdot \rVert$ is the magnitude of the gradient. The expression of the loss function for the ensemble model is:
\begin{equation}
        \mathcal{L}=\mathcal{L}_{ece} + \alpha \cdot \mathcal{L}_r + \beta \cdot \mathcal{L}_g
\end{equation}

\hspace*{\fill}
\hspace*{\fill}

\makeatletter
\newenvironment{breakablealgorithm}
{
	\begin{center}
		\refstepcounter{algorithm}
		\hrule height.8pt depth0pt \kern2pt
		\renewcommand{\caption}[2][\relax]{
			{\raggedright\textbf{\ALG@name~\thealgorithm} ##2\par}%
			\ifx\relax##1\relax 
			\addcontentsline{loa}{algorithm}{\protect\numberline{\thealgorithm}##2}%
			\else 
			\addcontentsline{loa}{algorithm}{\protect\numberline{\thealgorithm}##1}%
			\fi
			\kern2pt\hrule\kern2pt
		}
	}{
		\kern2pt\hrule\relax
	\end{center}
}
\makeatother

  

\subsection{Analysis of Ensemble Adversarial Defense}

\begin{figure}[h]
  \centering
  \includegraphics[width=.48\textwidth]{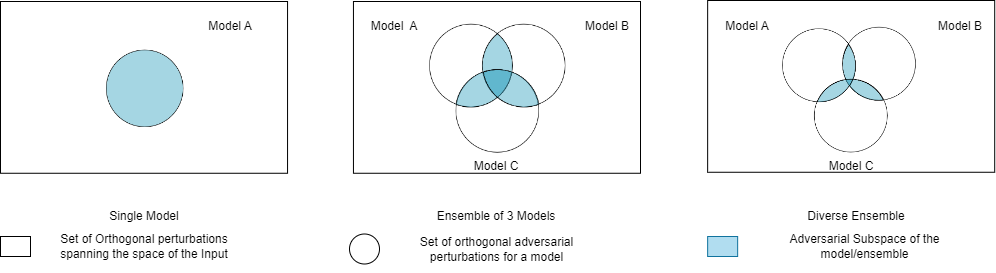}
  \caption{Adversarial Subspace}
  \label{Adversarial Subspace}
\end{figure}

\begin{figure}[h]
  \centering
  \includegraphics[width=.48\textwidth]{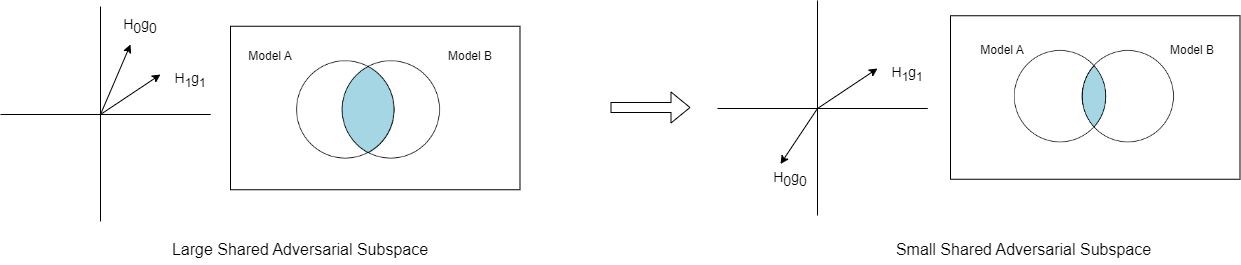}
  \caption{Heterogeneous Model Diversification}
  \label{Heterogeneous Model Diversification}
\end{figure}

The ensemble model has been validated as one of the effective means to enhance the robustness of a model. Figure~\ref{Adversarial Subspace} illustrates the fundamental concept of ensemble models, where an ensemble utilizes a collection of different models and aggregates their outputs for classification. The rectangle represents the space spanned by all possible orthogonal perturbations of input instances, while the circle denotes the subset of adversarial orthogonal perturbations that result in misclassifications by the model. The shaded region represents the perturbation subspace associated with the ensemble model.In the case of a single model, as depicted on the left side of Figure \ref{Heterogeneous Model Diversification}, any perturbation within the circle would lead to misclassification of the input image. However, for the scenario involving a collection of two or more models (depicted on the right and middle sides of Figure \ref{Heterogeneous Model Diversification}), successful adversarial attacks must occur within the shaded region. This implies that the attack must deceive all individual models in the ensemble. Therefore, fostering diversity among the models is an effective strategy for enhancing the ensemble model's adversarial robustness, as greater diversity among the submodels in the ensemble corresponds to less overlap between their respective adversarial subspaces.\par

To address the strong transferability of adversarial examples, we measure the effectiveness of our ensemble model from an attack transferability perspective. In a straightforward scenario, considering two models, the ideal situation is that adversarial examples generated under Model A are not misclassified by Model B. Therefore, even if one model makes incorrect predictions, the ensemble model would remain robust. Expanding this concept to an ensemble of $M$ models, we need to evaluate the transferability of each sub-model to all other sub-models. The final ensemble model fails only when the majority of the models in the ensemble are deceived. We focus on constructing multiple diverse sub-models and adopt a relatively simple decision strategy here, averaging the predictions of the M models. Thus, in our ensemble model, the decision result is incorrect only if $\left\lceil M/2 \right\rceil $ models are deceived.\par

To achieve this, we integrated multiple robust models with dispersed low curvature values to enhance the robustness of the ensemble model. The $L_g$ regularization term aims to control greater dispersion among the sub-models of the ensemble, thereby minimizing the overlap in the common adversarial subspace between the sub-models. As illustrated in Figure \ref{Heterogeneous Model Diversification}, by calculating the cosine angle values between the sub-models, denoted as Hg, we aim to obtain an ensemble model with a smaller dimension of the common adversarial subspace, as depicted on the right side of the figure.\par
The transfer success rate of adversarial attacks serves as a measure of similarity between two models. We can evaluate the transferability of adversarial examples by calculating the transfer success rate, indicating the attack effectiveness of adversarial examples across different models. A higher transfer success rate indicates better transferability of adversarial examples, where it easier to deceive multiple models. We utilize transferability success rate (TSR) as a metric to evaluate the defense capabilities of our ensemble model and compare it with those guided by first-order gradient-based methods.
\begin{equation}
    TSR=\frac{1}{M}\sum_{i=1}^{M}\sum_{j\neq i} 100*(1-\sum_{x \in \mathcal{D}}F_j(G_i(x)))
\end{equation}
where $G_i(x)$ represents the adversarial examples generated on the i-th sub-model, and $F_j(x)$ represents the probability of correct prediction on the j-th model. TSR is a metric that quantifies the ease of transfer between sub-models in an ensemble. A higher TSR indicates greater transferability.\par
\begin{table}[h]
\centering
\caption{TSR measured for an ensemble of 3 CNN models trained on CIFAR-100 and Tiny-ImageNet Datasets}\label{tab2}
\resizebox{0.5\textwidth}{!}{
\begin{tabular}{lcccccc}
\toprule%
& \multicolumn{3}{@{}c@{}}{CIFAR-100} & \multicolumn{3}{@{}c@{}}{Tiny-ImageNet} \\\cmidrule{2-4}\cmidrule{5-7}%
Para. & $\epsilon=0.01$ & $\epsilon=0.02$ & $\epsilon=0.04$ & $\epsilon=0.01$ & $\epsilon=0.02$ & $\epsilon=0.04$ \\
\midrule
Baseline  & 83.094  & 123.861 & 160.121 &59.982 & 108.494 &153.470 \\
DEG  & 101.545 &146.548   & 178.123 & 39.187 & 94.551 &162.048 \\
EDLCM  & \textbf{43.087}  & \textbf{92.026}  & \textbf{146.661} & \textbf{43.450}  & \textbf{91.849} &\textbf{148.601} \\
\bottomrule
\end{tabular}}
\end{table}
We computed the corresponding TSR on CIFAR-100 and Tiny-ImageNet datasets using PGD \cite{madry2017towards} as our attack method to measure transferability, where $\epsilon$ is the parameter controlling the magnitude of perturbation. As shown in Table 1, our method achieves the lowest TSR value. This indicates that EDLCM training can further reduce the transferability among sub-model members compared to gradient alignment. Furthermore, our regularization training method, compared to traditional ensemble defense methods based on first-order gradients, evidently enhances the robustness of ensemble models.

\section{Experiments}\label{sec4}
As discussed in this section, we conduct experiments on different datasets to validate the effectiveness of the proposed method. We first specify the experimental settings in Section A, then compare the robustness of our method with existing methods under white-box attacks \cite{kurakin2016adversarial} in Section B. In Section C, we further verify the robustness of our method under black-box attacks \cite{papernot2017practical}. Section D presents ablation experiments to explore the impact of $L_g$ and $L_r$.

\subsection{Experimental Setup}
\begin{itemize}
     \item \textbf{Evaluation Metric}: The accuracy of correctly classifying adversarial examples generated by the model serves as the evaluation criterion for assessing the robustness of our model. A higher accuracy(ACC) value indicates greater robustness.
    \item \textbf{Attack Methods}: We conduct untargeted adversarial attacks with four representative methods, namely FGSM \cite{goodfellow2014explaining}, BIM \cite{bim}, PGD \cite{madry2017towards}, and APGD \cite{croce2020reliable}. For fair comparisons, we follow the experimental settings described in Huang et al.\cite{huang2021adversarial} to set up the values $\epsilon$ for FGSM, BIM and PGD. The attack parameter in all tables is referred to as the attack strength for the four types mentioned above.
   \item \textbf{Ensemble Network Configuration}: The models were trained using the SGD optimizer \cite{robbins1951stochastic} with an initial learning rate of 0.02 with $\alpha=1$ and $\beta=0.01$ to control the penalty strength on $L_r$ and $L_g$, respectively. Under white-box attacks, we ensembled three ResNet-18 \cite{he2016deep} models on each dataset for convenient comparison. We used VGG-16 \cite{simonyan2014very} and ResNet-34 \cite{he2016deep} to evaluate our regularization methods under different types black-box attacks. 
\end{itemize}

\begin{table*}[ht]
\centering
\caption{Prediction accuracy (\%) on adversarial examples generated by white-box attacks}
\renewcommand\arraystretch{1.5}
\resizebox{1\linewidth}{!}{\begin{tabular}{ccccccccccccccccccc}
  \hline
         \multirow{2}*{\makecell[c]{Attacks}} & \multicolumn{6}{c}{\makecell[c]{CIFAR10}} & \multicolumn{6}{c}{\makecell[c]{CIFAR100}}& \multicolumn{6}{c}{\makecell[c]{Tiny-Imagenet}}\\
        \cline{2-19}
         & Para. & Base & DEG & ADP& PDD& EDLCM& Para.& Base &  DEG &  ADP& PDD&EDLCM& Para. &Base& DEG &  ADP&  PDD &EDLCM\\
        \hline
         \hline
           No Attack & -  &95.950& 95.630 &95.850 &95.730 &94.730&- & 80.460& 80.570  &80.650 &79.900 &76.650& -& 69.710& 69.430& 68.050&  64.510&67.810 \\
           \hline\hline 
         \multirow{2}*{PGD}   & 0.005        & 70.641 & 36.673 & 77.413 & 80.111 & \textbf{91.553}
 & 0.01         & 17.226 & 13.156 & 26.646 & 41.589 & \textbf{64.549}& 0.01         & 28.203 & 23.855 & 23.401 & 37.390 & \textbf{50.258} 
 \\
           & 0.01         & 36.196 & 6.766  & 59.113 & 60.086 & \textbf{79.289}
  & 0.02         & 3.144  & 2.321  & 10.192 & 21.840 & \textbf{34.405}& 0.02         & 7.216  & 6.598  & 3.956  & 21.795 & \textbf{22.519}  \\
        \hline
          \multirow{2}*{BIM}  & 0.005        & 62.011 & 33.640 & 73.438 & 75.159 & \textbf{88.179}& 0.01         & 12.839 & 11.208 & 23.794 & 37.647 & \textbf{55.007}& 0.01         & 20.958 & 19.015 & 14.475 & 33.328 & \textbf{41.203}  \\
          
         & 0.01         & 28.463 & 7.268  & 54.846 & 53.881 & \textbf{71.127}
  & 0.02         & 2.361  & 2.110  & 11.035 & 21.314 & \textbf{23.895}& 0.02         & 4.404  & 4.624  & 2.145  & 19.780 & 15.042 
  \\
        \hline
           \multirow{2}*{FGSM}  & 0.005        & 74.987 & 74.736 & 82.556 & 82.994 & \textbf{89.718} & 0.01         & 38.976 & 42.472 & 46.770 & 53.692 & \textbf{60.627} & 0.01         & 41.084 & 50.072 & 27.299 & 53.992 & 50.730
 \\
         & 0.01         & 59.541 & 45.728 & 73.333 & 71.472 & \textbf{75.525}& 0.02         & 23.726 & 20.442 & 36.875 & 36.696 & 34.353 & 0.02         & 22.967 & 34.386 & 13.051 & 39.653 & 27.946   \\
        \hline
         \multirow{2}*{APGD}
                  & 0.005        & 32.371 & 1.129  & 43.349 & 56.450 & \textbf{85.437}& 0.01         & 1.666  & 0.348  & 2.976  & 14.708 & \textbf{46.680} & 0.01         & 5.222  & 2.377  & 5.642  & 17.860 & \textbf{27.680} 
\\  
                 & 0.01  & 2.731  & 0      & 8.534  & 19.263 & \textbf{59.127} & 0.02  & 0.087  & 0      & 0.112  & 3.254  & \textbf{11.359} & 0.02  & 0.172  & 0.058  & 0.264  & 7.513  & 3.333  \\
                 \hline

\end{tabular}
  }
\end{table*}
\begin{figure}[ht]
    \centering
    \includegraphics[width=.42\textwidth]{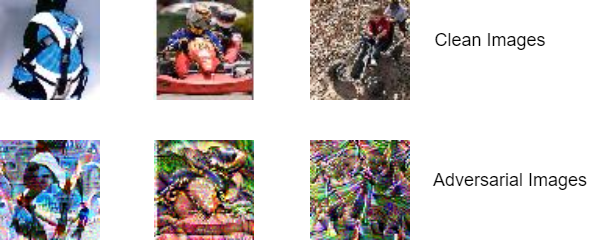}
    \caption{Adversarial examples generated on TinyImageNet under BIM}
    \label{adversarial_examples}
\end{figure}
\begin{figure*}
  \centering
  \subfloat[FGSM]{\includegraphics[width=.25\textwidth]{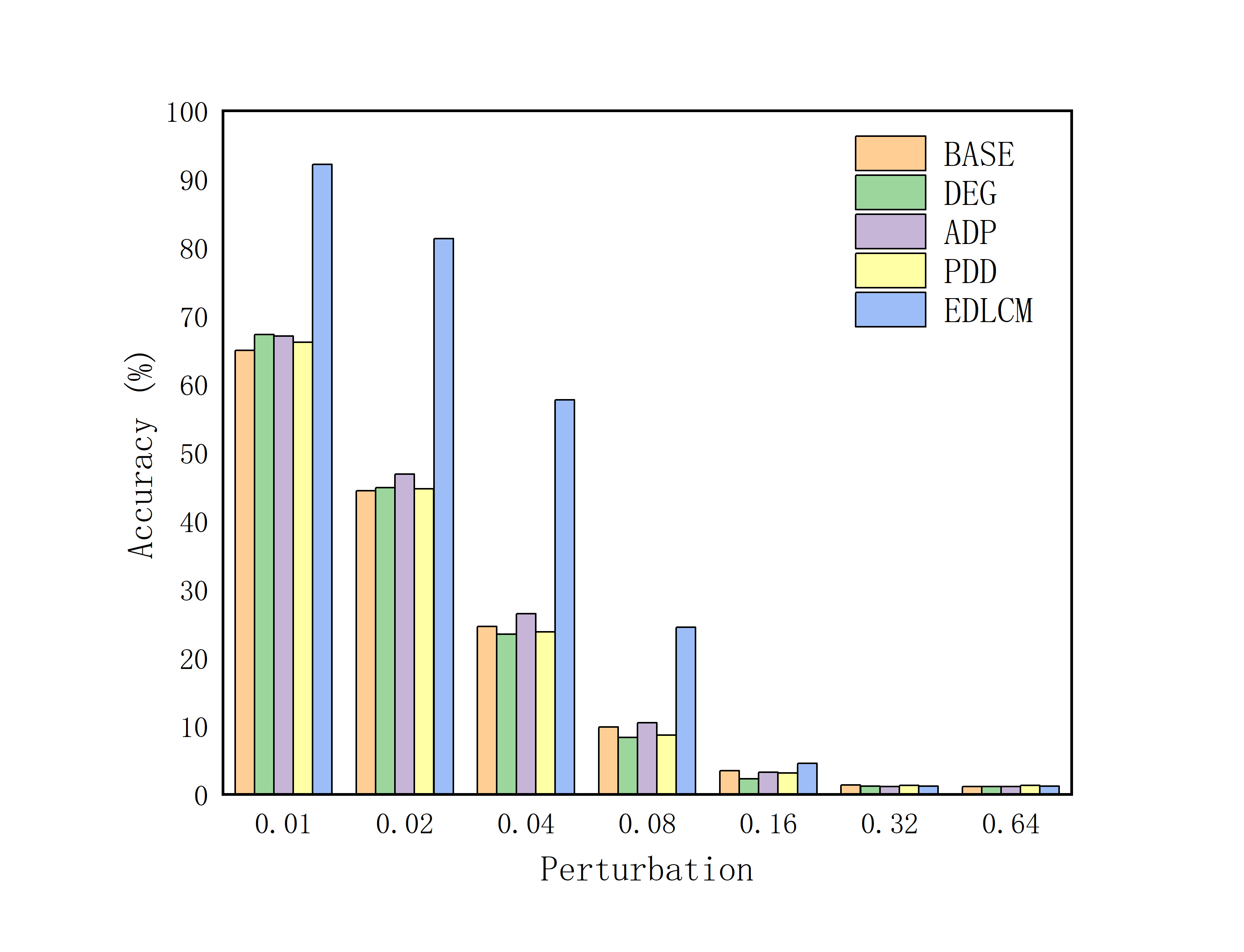}}
  \subfloat[PGD]{\includegraphics[width=.25\textwidth]{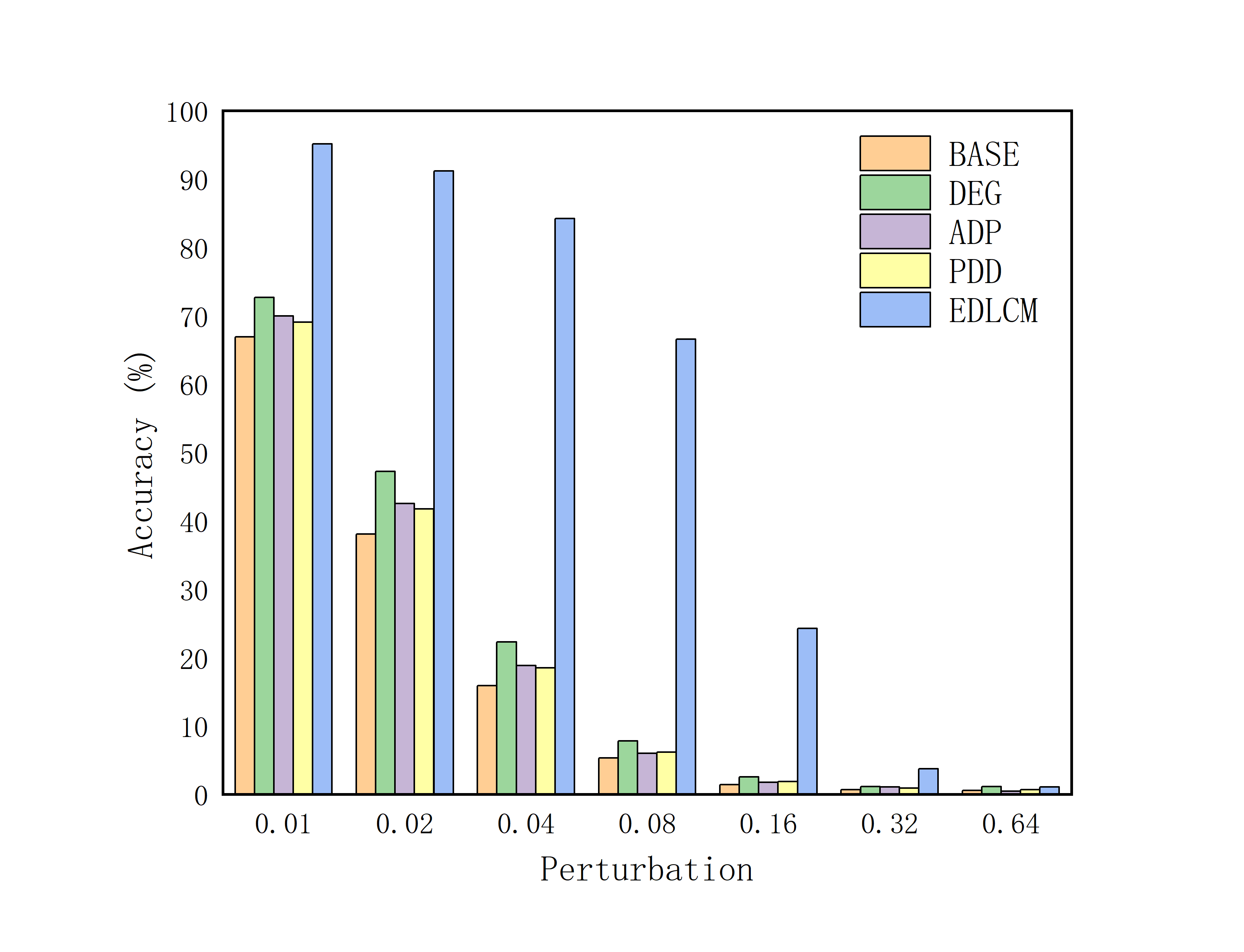}}
  \subfloat[BIM]{\includegraphics[width=.25\textwidth]{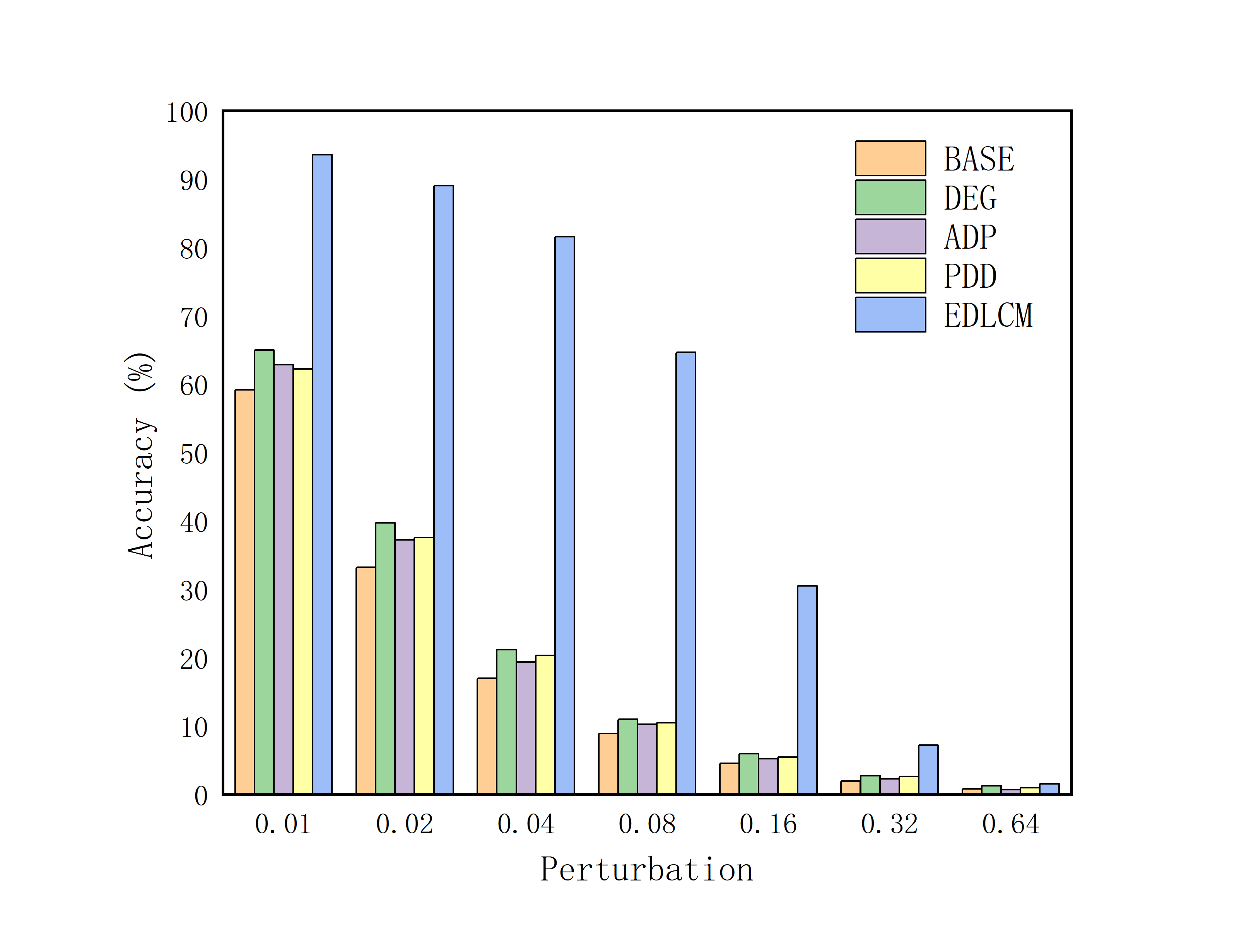}}
  \subfloat[APGD]{\includegraphics[width=.25\textwidth]{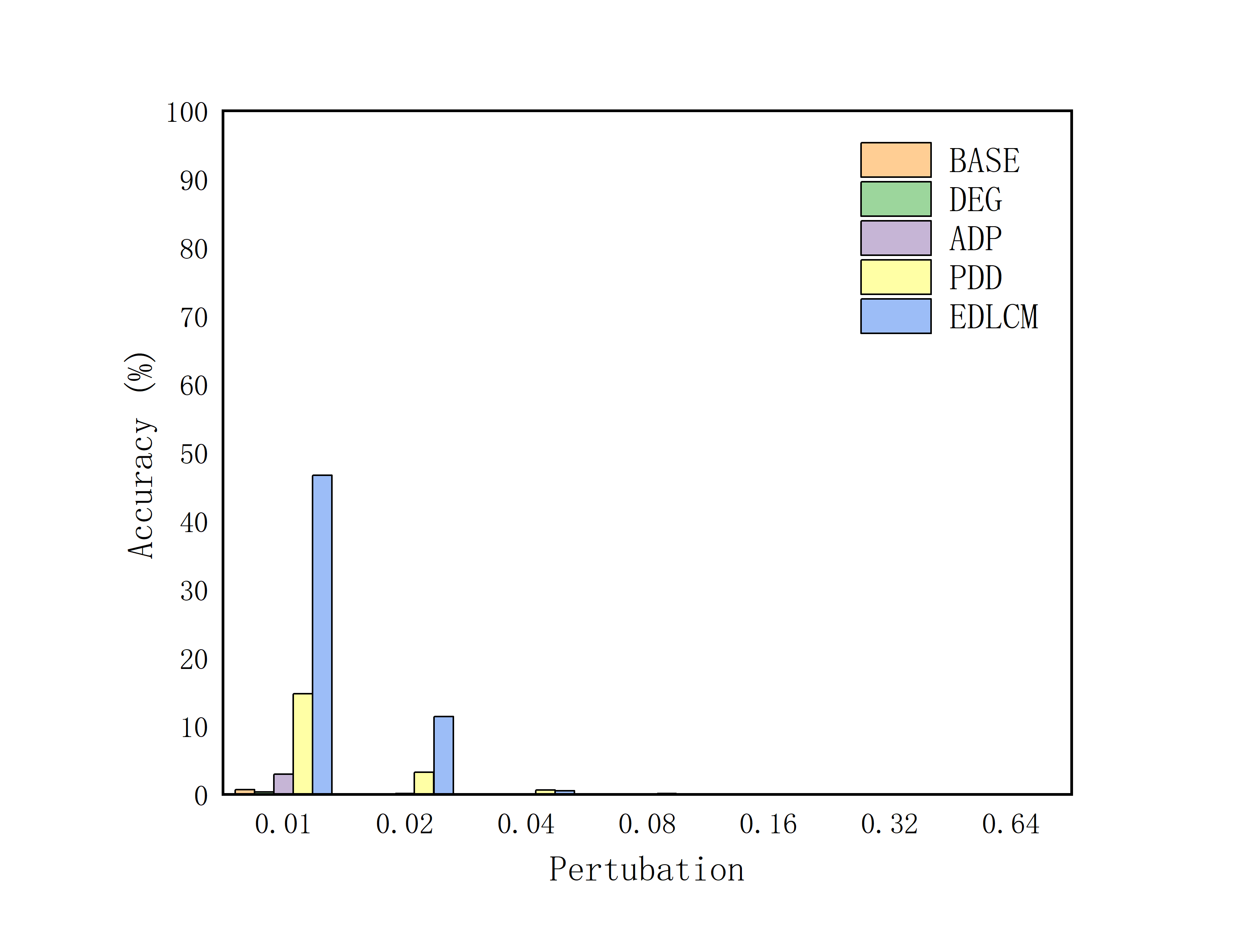}}
  \caption{Comparative Analysis of Defense Performance for Various Ensemble Defense Strategies Against Type 1 Black-Box Attacks on  CIFAR-100 }
  \label{blackattack}
\end{figure*}

\begin{figure*}
  \centering
  \subfloat[FGSM]{\includegraphics[width=.25\textwidth]{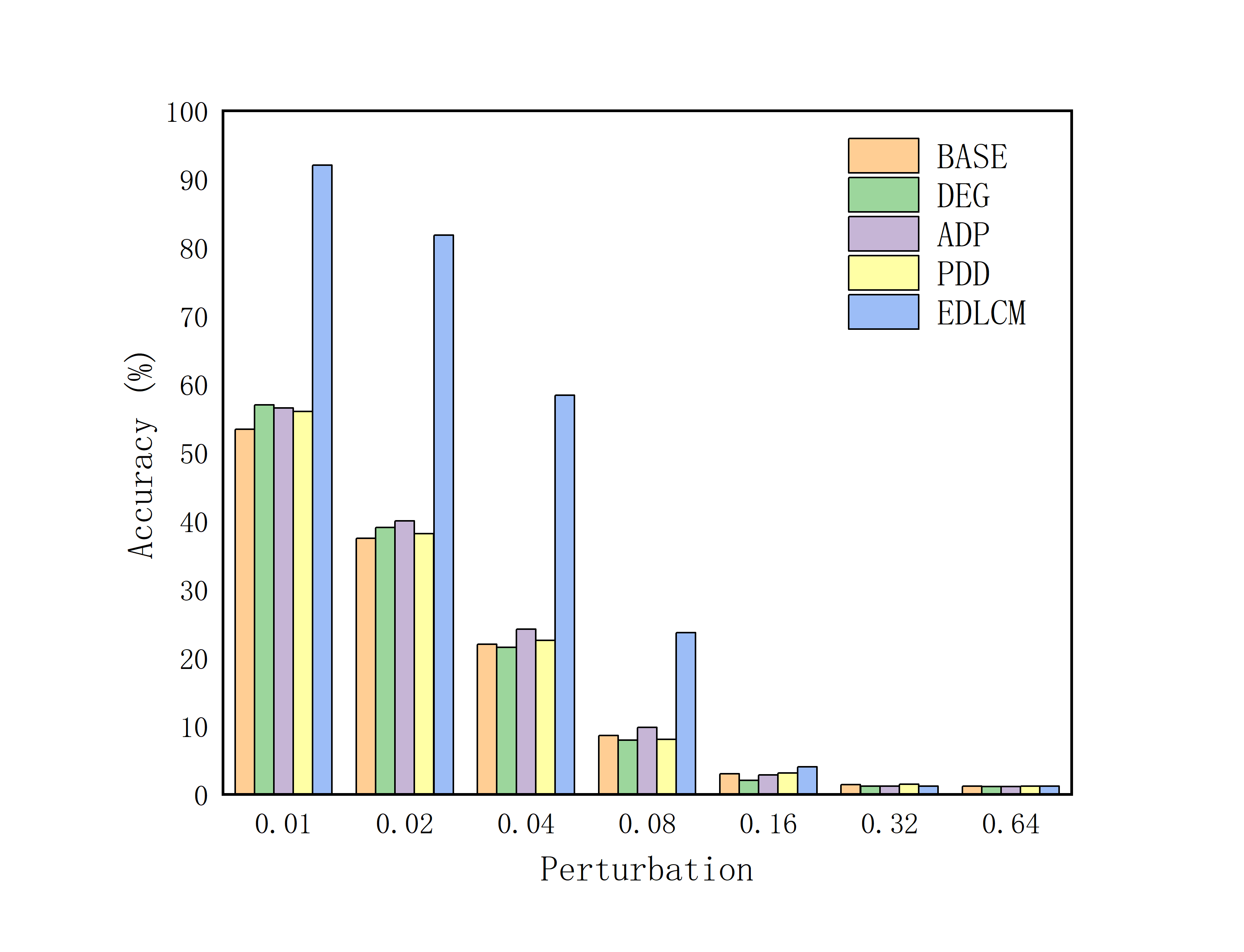}}
  \subfloat[PGD]{\includegraphics[width=.25\textwidth]{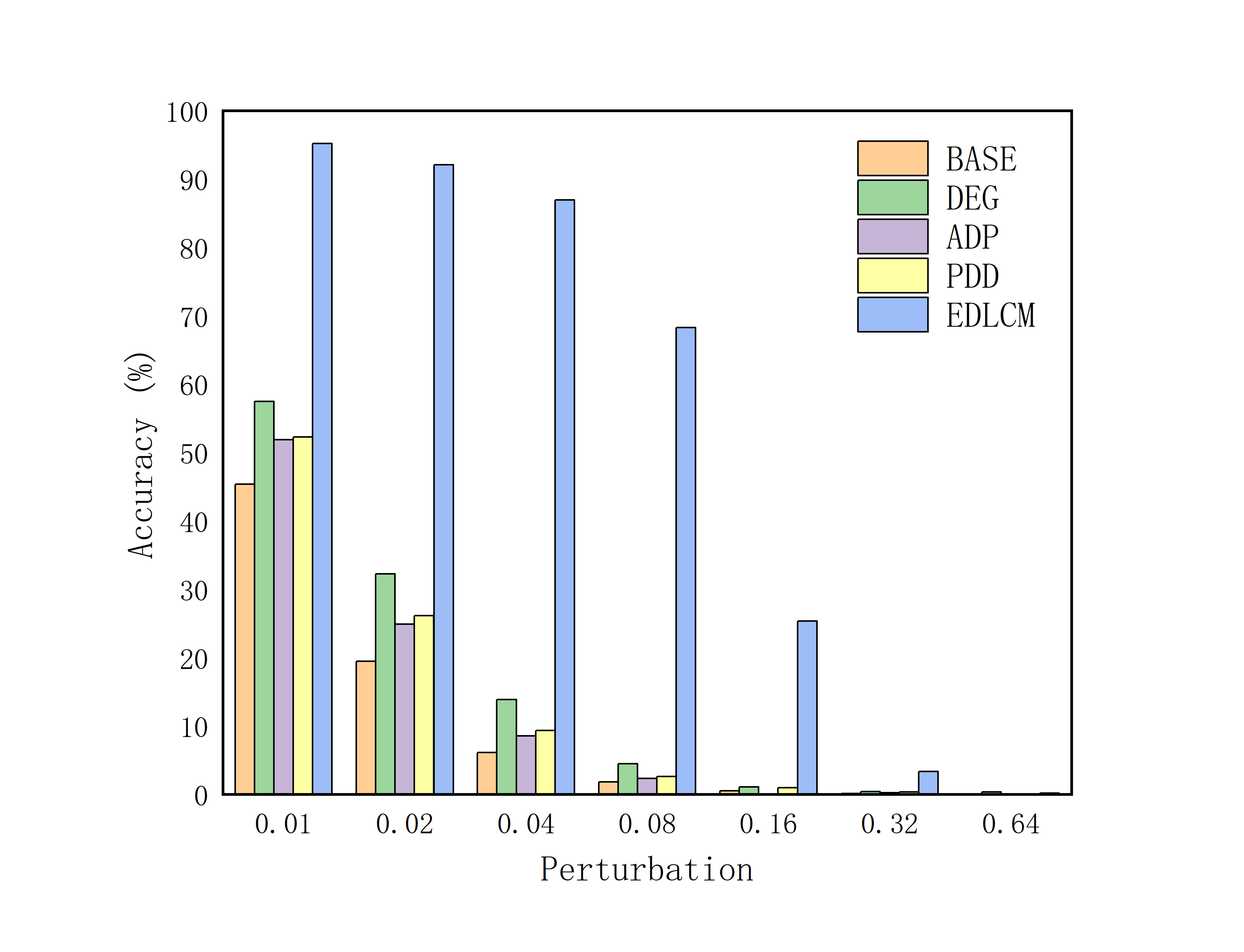}}
  \subfloat[BIM]{\includegraphics[width=.25\textwidth]{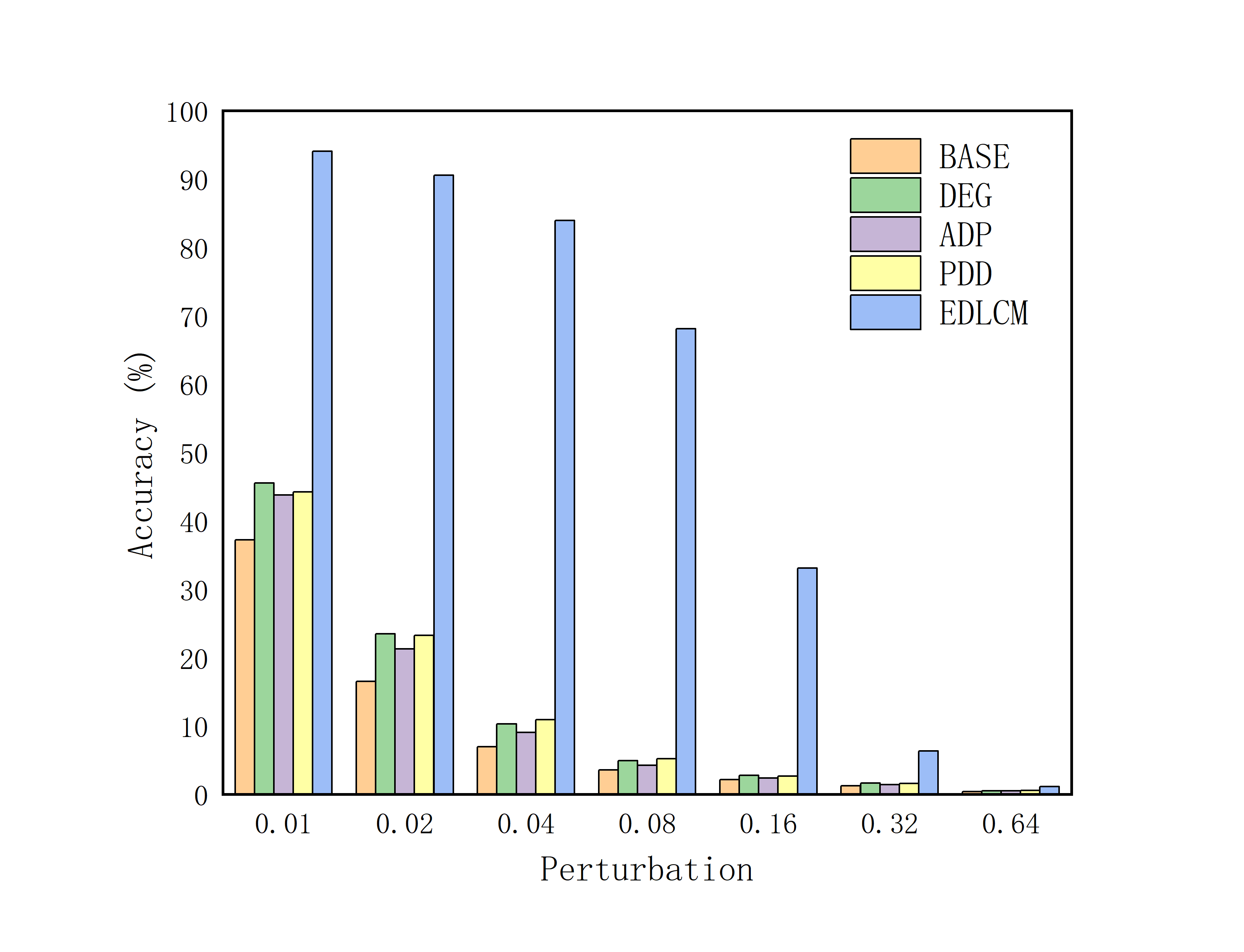}}
  \subfloat[APGD]{\includegraphics[width=.25\textwidth]{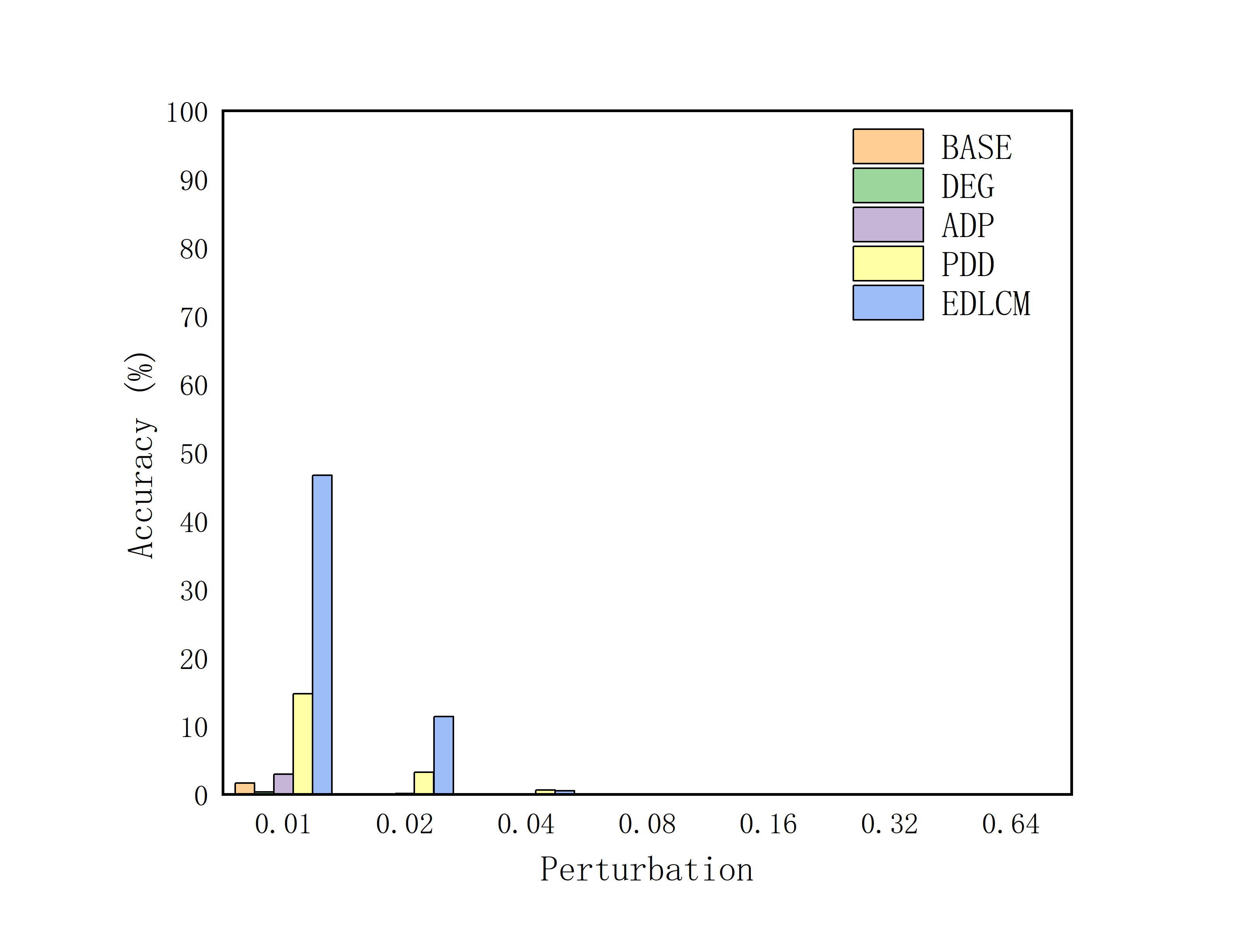}}
  \caption{Comparative Analysis of Defense Performance for Various Ensemble Defense Strategies Against Type 2 Black-Box Attacks on  CIFAR-100 }
  \label{blackattack2}
\end{figure*}
\subsection{Results under White-box Attacks}
White-box attacks assume that the attacker possesses complete access to and knowledge of the internal information of the target model, including its structure, parameters, input, output, and other relevant information.\par

\begin{figure}[ht]

  \subfloat[CIFAER-10]{\includegraphics[width=.53\linewidth]{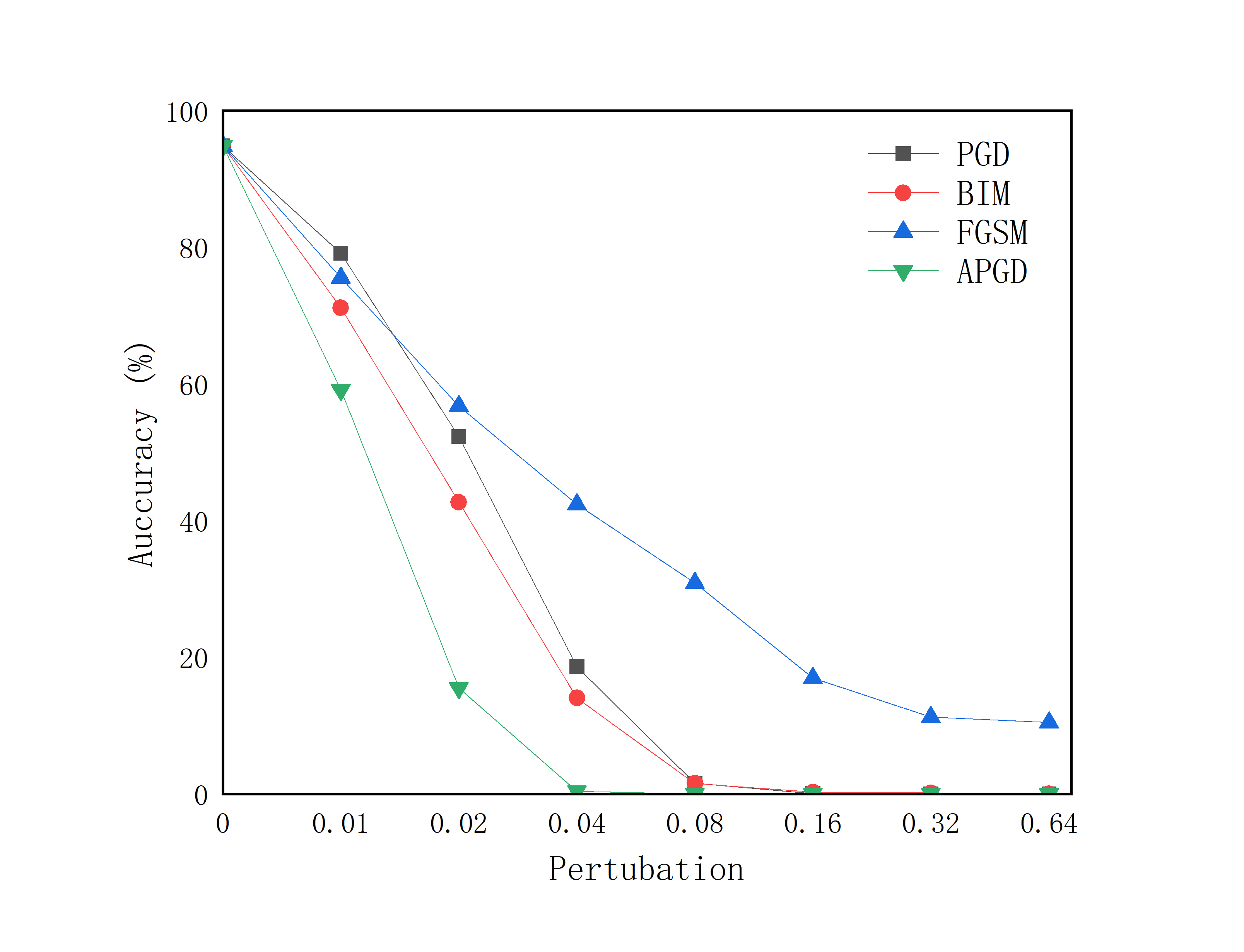}}
  \subfloat[CIFAR-100]{\includegraphics[width=.53\linewidth]{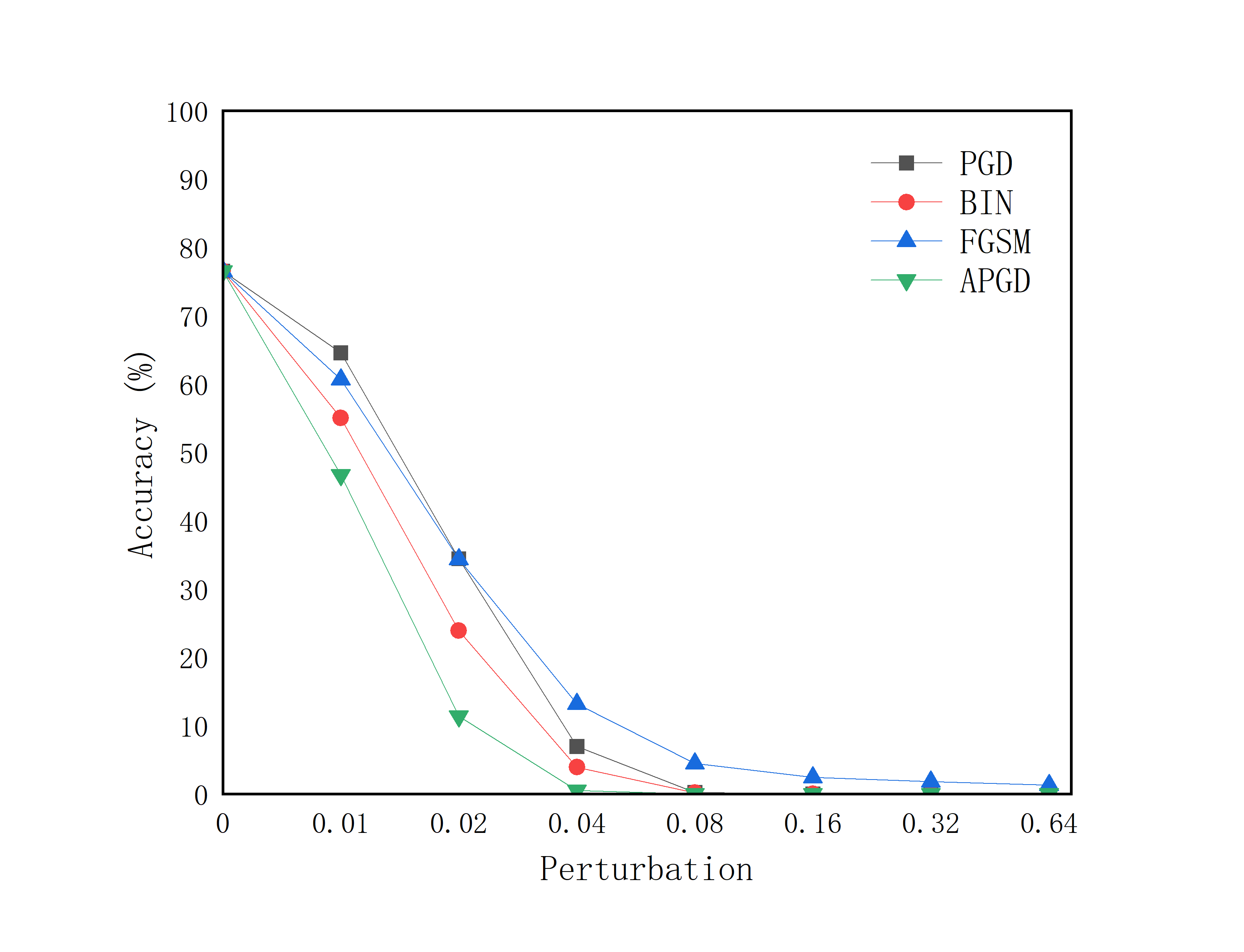}}
  \caption{The Robustness of EDLCM under Different Perturbation Intensities}
  \label{edlcm}
\end{figure}
As depicted in the tables below, our method significantly improves adversarial robustness while maintaining high accuracy on normal examples. The experiments described in Table 2 illustrate the effectiveness of our method. Given the same level of attack strength, our approach demonstrates significant improvements over ADP \cite{pang2019improving}, PDD \cite{huang2021adversarial}, and DEG \cite{huang2021adversarial} across various adversarial attack methods. Taking CIFAR-100 as an example, under a PGD (0.01) attack, the recognition accuracy of the baseline and DEG is less than 20\%, the ADP method only achieves an accuracy of only 26.65\%, and the PDD method attains an accuracy of only 41.59\%. However, our method achieves a recognition accuracy of around 65\%. Similar experimental results were also observed on CIFAR10 and Tiny-ImageNet, where our method outperforms all other tested methods under PGD and BIM attacks.\par
Though the proposed method does not perform as well as as the PDD method under FGSM attacks, it is still comparable and significantly more accurate than ADP or DEG. For instance, under an FGSM attack with a perturbation of 0.01, our method exhibits an improvement of nearly 50\% compared to the ADP method. This indicates that our ensemble defense method outperforms existing ensemble strategies and maintains high accuracy even under high-intensity adversarial attacks.

\subsection{Results under Black-box Attacks}
Black-box attacks assume that the attacker has limited knowledge of the target model. In this scenario, the attacker lacks access to the internal information of the model, such as its architecture and parameters. Black-box attacks can be further categorized into two types: one where the attacker has no knowledge of the model's defense mechanisms and network architecture, and another where the attacker possesses partial information about the model but lacks details about the specific implementation of the defense, such as using a similar network architecture or understanding the model's defense mechanisms. We separately  tested for defense effectiveness under two types of black-box attacks and compared it with existing ensemble defense methods. In the first category of pure black-box attacks, we employed VGG16 as a surrogate model for black-box attacks, which possesses a different architecture from the target model. For the second type of black-box attacks, we conducted black-box attack tests using ResNet34, a model with a structure similar to the target model.\par

In the first category of black-box attacks, as illustrated in Figure \ref{blackattack}, we observed that existing ensemble defense methods perform similarly under different types of black-box attacks. Defending against black-box attacks becomes more challenging as the perturbation level increases, as evidenced by the fluctuation of the accuracy of the ensemble model at around 38\% under BIM (0.02). The improved optimization-based ensemble algorithms DEG, ADP, PDD show robustness enhancements within a fluctuation of 5 percentage points. This suggests that adversarial samples generated by the surrogate model can still effectively transfer to existing ensemble defense methods like DEG.

However, for our ensemble model, high classification accuracy is retained under the same black-box attack. Under PGD attacks with a perturbation strength of 0.02, we achieved an almost 50\% improvement in recognition accuracy. Similar experimental results were verified with BIM and FGSM attacks. Even under the latest attack method, APGD, existing ensemble strategies do not effectively defend against the adversarial examples generated while our method retains a certain level of recognition accuracy.This indicates a significant improvement in the adversarial robustness of our approach.\par

Similar experimental results were validated under the second category of black-box attacks, as depicted in Figure \ref{blackattack2}. When subjected to black-box attacks using similar network structures, existing ensemble defense methods such as PDD experienced varying degrees of degradation in robustness. Under various intensities of black-box attacks, the robustness performance of ensemble models constructed by methods like PDD was comparable, with marginal improvement compared to BASE. This indicates a bottleneck in existing ensemble defense strategies when confronted with black-box attacks.

However, under our ensemble defense strategy, our ensemble model maintains a high level of robustness. Moreover, it demonstrates a nearly 50\% improvement in performance compared to existing ensemble models, confirming the effectiveness of our proposed regularization training-based model heterogeneous ensemble strategy under various black-box attacks.

\subsection{Ablation Tests}
\begin{figure*}[ht]
  \centering
  \subfloat[Single Model]{\includegraphics[width=.33\textwidth]{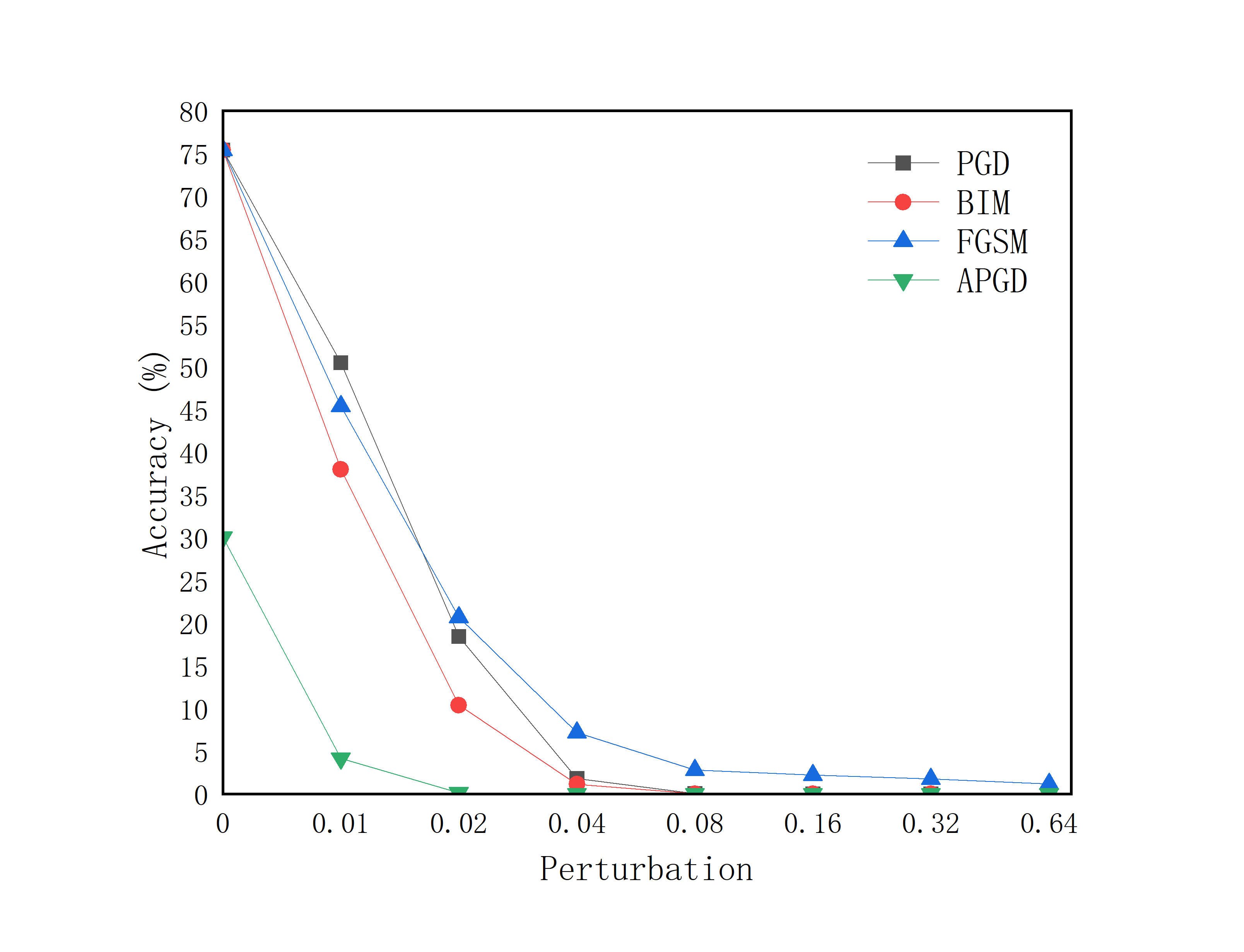}}
  \subfloat[Three Models]{\includegraphics[width=.33\textwidth]{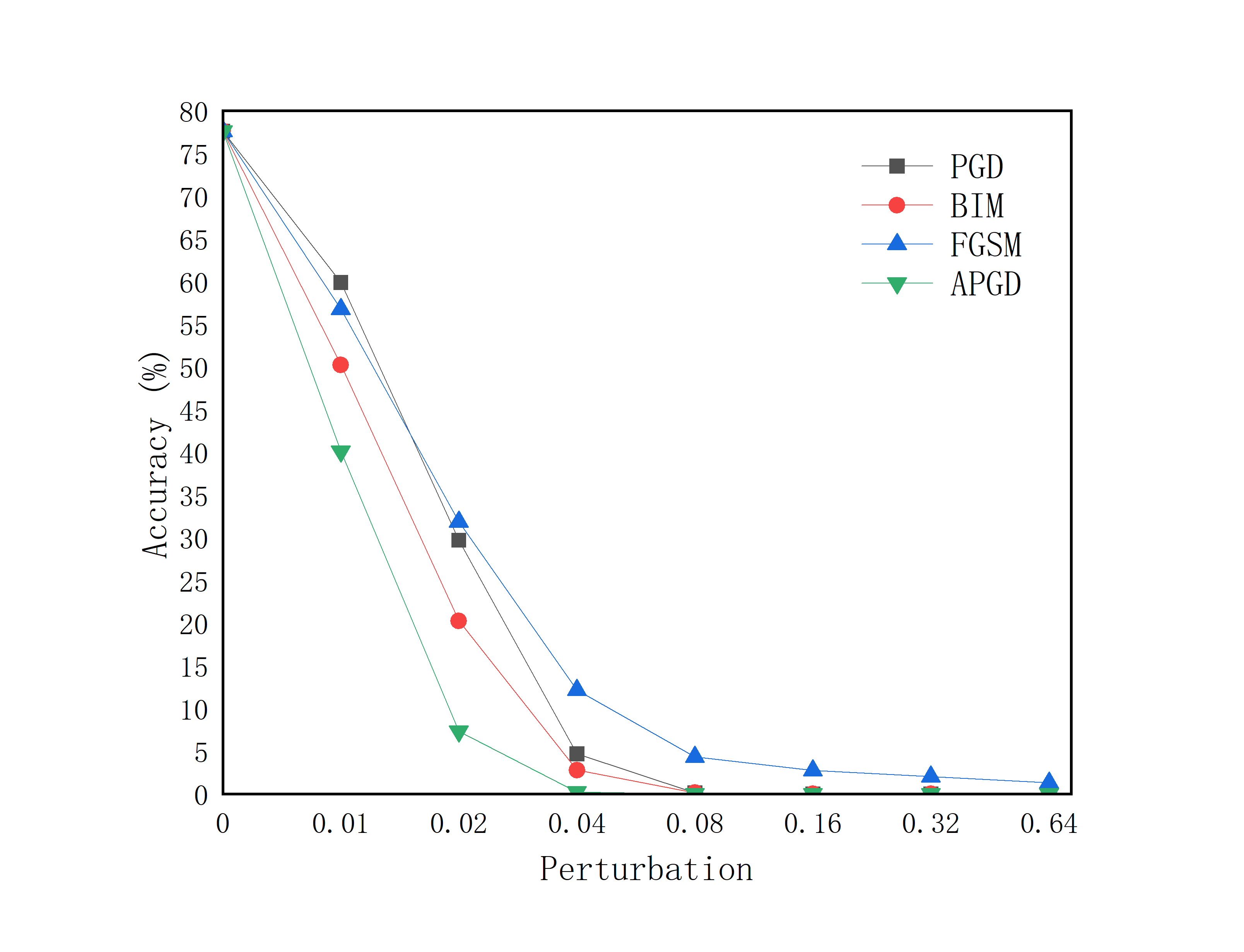}}
  \subfloat[Five Models]{\includegraphics[width=.33\textwidth]{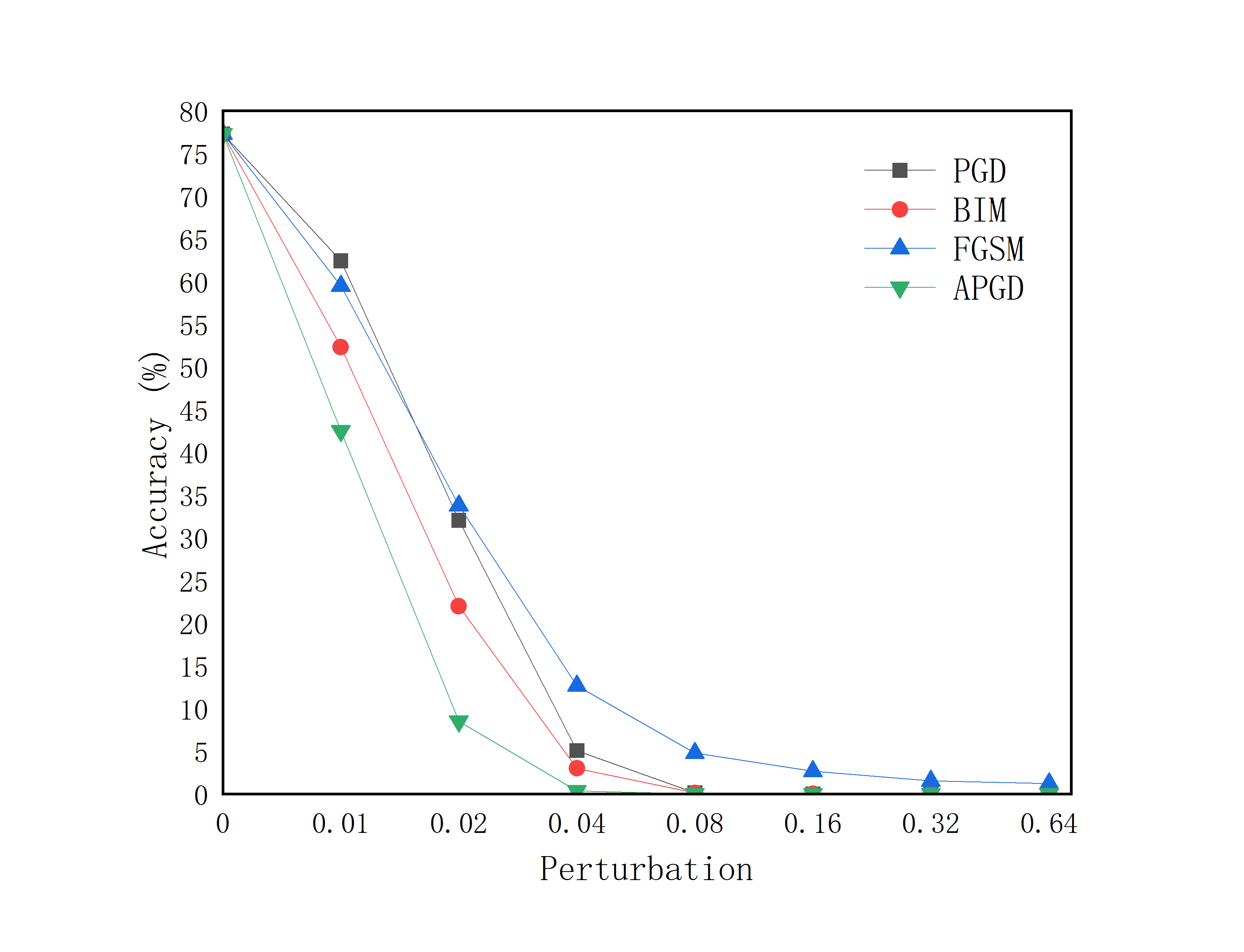}}
  \caption{Comparison of prediction accuracy (\%) on adversarial examples generated by white-box attacks on CIFAR-100 of $L_r$ between M networks}
  \label{cifar100models}
\end{figure*}
We also examined the individual effects of the two regularization terms in our method on ensemble defense.We initially assessed the impact of the $L_r$ regularization term on the robustness improvement of ensemble model members. Relevant experiments were carried out on CIFAR-100, confirming the effectiveness of integrating system defense through the ensemble of low-curvature and highly robust models. As illustrated in Figures \ref{cifar100models}, we evaluated the robustness performance of the ensemble model when the number of ensemble model members (M) varied (M=1, 3, 5). We observed that the defense effectiveness of the ensemble model reached its optimal level when the number of ensemble members was 5. However, as the number of ensemble model members increased, the improvement in the ensemble model's defense became less significant. This indicates that enhancing robustness solely by increasing the number of individual models in the ensemble has limitations. It is crucial to consider the diversity among ensemble model members, namely, the degree of heterogeneity.\par

Secondly, we validated the impact of the $L_g$ regularization term on the heterogeneity of the ensemble model. We compared the defensive performance of the ensemble models trained with BASE, DEG, and solely the $L_g$ regularization term on CIFAR-100. The experimental results, as shown in Table \ref{Lg}, indicate that by constructing diverse base sub-models through cosine angle computation between models, $L_g$ exhibits some improvement potential compared to traditional first-order gradient alignment methods. This suggests that second-order gradients can integrate more diverse sub-model members, thereby enhancing the comprehensive defense capability of the ensemble model.\par
\begin{table}[ht]
\centering
\caption{Comparison of prediction accuracy (\%) on adversarial examples generated by white-box attacks on CIFAR-100 between different ensemble defenses. These defense methods compute the cosine similarity among model members to promote diversity within the ensemble.}
\renewcommand\arraystretch{0.6}
\begin{tabular}{cllll}
\toprule
\multirow{2}{*}{Attack} & \multicolumn{4}{c}{CIFAR-100}                                                                                   \\ \cmidrule{2-5} 
                        & \multicolumn{1}{l}{Para.} & \multicolumn{1}{c}{Base}   & \multicolumn{1}{c}{DEG}    & \multicolumn{1}{c}{only $L_g$} \\ \midrule

\multirow{2}{*}{PGD}    & \multicolumn{1}{c}{0.01} & \multicolumn{1}{l}{17.226} & \multicolumn{1}{l}{13.156} & \textbf{21.633}          \\ \cmidrule{2-5} 
                        & \multicolumn{1}{c}{0.02} & \multicolumn{1}{l}{3.144}  & \multicolumn{1}{l}{2.321}  & \textbf{3.812}           \\ \midrule
\multirow{2}{*}{BIM}    & \multicolumn{1}{c}{0.01} & \multicolumn{1}{l}{12.839} & \multicolumn{1}{l}{11.208} & \textbf{16.102}          \\ \cmidrule{2-5} 
                        & \multicolumn{1}{c}{0.02} & \multicolumn{1}{l}{2.361}  & \multicolumn{1}{l}{2.110}  & \textbf{2.972}                    \\ \midrule
\multirow{2}{*}{FGSM}   & \multicolumn{1}{c}{0.01} & \multicolumn{1}{l}{38.976} & \multicolumn{1}{l}{42.472} & 39.842                   \\ \cmidrule{2-5} 
                        & \multicolumn{1}{c}{0.02} & \multicolumn{1}{l}{23.726} & \multicolumn{1}{l}{20.442} & \textbf{24.944}          \\ \midrule
\multirow{2}{*}{APGD}   & \multicolumn{1}{c}{0.01} & \multicolumn{1}{l}{1.666} & \multicolumn{1}{l}{0.348} &      \textbf{2.257}             \\ \cmidrule{2-5} 
                        & \multicolumn{1}{c}{0.02} & \multicolumn{1}{l}{0.087} & \multicolumn{1}{c}{0} & \textbf{0.088}          \\ \bottomrule

\end{tabular}
\footnotetext{Best performance is highlighted in bold}
\label{Lg}
\end{table}

Finally, we compared the impact of the two regularization terms, $L_g$ and $L_r$, on ensemble defense in our approach. The experimental results, as presented in Table \ref{LgLr}, demonstrate that the regularization training with $L_g$ and $L_r$ effectively strengthens the robustness of the ensemble model, improving its defensive performance. For instance, on the CIFAR-100 dataset, under PGD (0.01) attack intensity, the ensemble model trained with $L_g$ alone achieved only 21.63\% robustness, while the $L_r$ trained ensemble model reached a robustness of 60.031\%. Combining $L_g$ and $L_r$ in the ensemble model further enhanced robustness to 64.549\%. $L_r+L_g$ achieves the best performance. This indicates that $L_g$ and $L_r$ can complement each other to some extent, leading to a further reduction in the transferability of adversarial attacks among sub-models and enhancing the overall robustness of the model.\par
\begin{table}[!ht]
\centering
\caption{Comparison of prediction accuracy (\%) on adversarial examples generated by white-box attacks on CIFAR-100 between different regularization terms}
\renewcommand\arraystretch{0.6}
\begin{tabular}{clllll}
\toprule
\multirow{2}{*}{Attack} & \multicolumn{5}{c}{CIFAR-100}                                                                                                        \\ \cmidrule{2-6} 
                        & \multicolumn{1}{l}{Para.} & \multicolumn{1}{c}{Base}   & \multicolumn{1}{c}{only $L_g$}    & \multicolumn{1}{c}{only $L_r$}    & $L_r+L_g$         \\ \midrule
\multirow{2}{*}{PGD}    & \multicolumn{1}{c}{0.01} & \multicolumn{1}{l}{17.226} & \multicolumn{1}{l}{21.633} & \multicolumn{1}{l}{60.031} & \textbf{64.549} \\ \cmidrule{2-6} 
                        & \multicolumn{1}{c}{0.02} & \multicolumn{1}{l}{3.144}  & \multicolumn{1}{l}{3.812}  & \multicolumn{1}{l}{29.790} & \textbf{34.405} \\ \midrule
\multirow{2}{*}{BIM}    & \multicolumn{1}{c}{0.01} & \multicolumn{1}{l}{12.839} & \multicolumn{1}{l}{16.102} & \multicolumn{1}{l}{50.238} & \textbf{55.007} \\ \cmidrule{2-6} 
                        & \multicolumn{1}{c}{0.02} & \multicolumn{1}{l}{2.361}  & \multicolumn{1}{l}{2.972}  & \multicolumn{1}{l}{20.268} & \textbf{23.895} \\ \midrule
\multirow{2}{*}{FGSM} & \multicolumn{1}{c}{0.01} & \multicolumn{1}{l}{38.976} & \multicolumn{1}{l}{39.842} & \multicolumn{1}{l}{56.771} & \textbf{60.627} \\ \cmidrule{2-6} 
                        & \multicolumn{1}{c}{0.02} & \multicolumn{1}{l}{23.726} & \multicolumn{1}{l}{24.944} & \multicolumn{1}{l}{31.852} & \textbf{34.353} \\ \midrule
\multirow{2}{*}{APGD} & \multicolumn{1}{c}{0.01} & \multicolumn{1}{l}{1.666} & \multicolumn{1}{l}{2.257} & \multicolumn{1}{l}{40.085} & \textbf{46.680} \\ \cmidrule{2-6} 
                        & \multicolumn{1}{c}{0.02} & \multicolumn{1}{l}{0.087} & \multicolumn{1}{l}{0.088} & \multicolumn{1}{l}{7.319} & \textbf{11.359} \\ \bottomrule                   
\end{tabular}
\footnotetext{Best performance is highlighted in bold}
\label{LgLr}
\end{table}
However, our method also has certain limitations in terms of time efficiency. We tested the training time per epoch with a mini-batch size of 64. Taking CIFAR-100 as an example, when integrating three ResNet-18 models, it takes 23s/epoch for the baseline, 65s/epoch for ADP, 46s/epoch for PDD, 259s/epoch for DEG, and 596s/epoch for EDLCM on Tesla A100. Optimizing the expensive time cost of the proposed method is an important future research direction.

\section{Conclusion}

This paper introduces a novel regularized ensemble defense method. In contrast to previous research predominantly centering on first-order gradients, we emphasize the impact of second-order gradients on the transferability of adversarial attacks. We demonstrate that reducing the curvature values of all sub-model, coupled with increasing the dispersion among them, yields a more diverse ensemble model with better adversarial robustness. While our method outperforms existing ensemble defense methods, it does have some noteworthy limitations. The computation of second-order gradients increases the time cost of our training process, rendering it less efficient than traditional approaches. Our future research objectives include optimizing the computational efficiency of gradient information and devising new techniques to approximate the Hessian matrix. Additionally, while this study primarily addresses the construction of multiple decentralized member models, the decision-making process based on member outputs is simplified through average summation. Investigating advanced decision-making strategies based on member outputs is another important goal of our continued research.


\section*{Acknowledgment}
This work is supported by the National Key Research and Development Program of China under Grant No. 2022YFB4500900.
\bibliographystyle{unsrt}

\bibliography{paper}

\end{document}